\pdfoutput=1
\PassOptionsToPackage{table}{xcolor}
\documentclass[letterpaper]{article} %
\usepackage[preprint]{aaai2027_arxiv}
\usepackage[hyphens]{url}  %
\usepackage{graphicx} %
\usepackage{natbib}  %
\usepackage{caption} %
\usepackage{algorithm}
\usepackage{algorithmic}
\usepackage{multirow}
\usepackage{amsmath}
\usepackage{amsfonts}

\usepackage{tabularx}
\usepackage{array}
\usepackage{booktabs}
\newcolumntype{Y}{>{\centering\arraybackslash}X}

\usepackage{newfloat}
\usepackage{listings}
\DeclareCaptionStyle{ruled}{labelfont=normalfont,labelsep=colon,strut=off} %
\floatstyle{ruled}
\newfloat{listing}{tb}{lst}{}
\floatname{listing}{Listing}

\usepackage{booktabs}
\usepackage{xcolor}
\definecolor{mygray}{HTML}{e5e5e5}

\usepackage{amssymb}
\usepackage{enumitem}
\definecolor{prompttitle}{HTML}{569fd8}
\definecolor{promptbackground}{HTML}{dbe7f5}
\usepackage[most,skins,theorems]{tcolorbox}
\tcbset{
  aibox/.style={
    width=\linewidth,
    top=8pt,
    bottom=4pt,
    colback=promptbackground,
    colframe=white,
    colbacktitle=prompttitle,
    enhanced,
    center,
    attach boxed title to top left={yshift=-0.1in,xshift=0.15in},
    boxed title style={boxrule=0pt,colframe=white,},
  }
}
\newtcolorbox{AIbox}[2][]{aibox,title=#2,#1}

\newcommand{\ModelName}{FALCA}

\title{\ModelName{}: Falsification-Aware Reinforcement Learning with Hierarchical Credit Assignment for Long-Horizon Deep Search Agents}
\author{
    Zijun Dou\textsuperscript{\rm 1,\rm 2}\equalcontrib,
    Kun Luo\textsuperscript{\rm 2,\rm 3}\equalcontrib,
    Chaofan Li\textsuperscript{\rm 1},
    Shuqi Lu\textsuperscript{\rm 1},
    Zheng Liu\textsuperscript{\rm 1}\corresponding
}
\affiliations{
    \Large
    \textsuperscript{\rm 1} BAAI, \ \ \ \
    \textsuperscript{\rm 2} THU, \ \ \ \
    \textsuperscript{\rm 3} CASIA\\
    \texttt{\{royz.dou,luokun695,zhengliu1026\}@gmail.com}
}

\definecolor{citecolor}{HTML}{387FBC}
\usepackage[colorlinks=true,linkcolor=citecolor,citecolor=citecolor,urlcolor=citecolor]{hyperref}

\begin{document}

\maketitle
{\renewcommand\thefootnote{}\footnotetext{Homepage: \url{https://douzj-llm.github.io/FALCA/}}}

\begin{abstract}
Long-horizon deep search requires LLM agents to iteratively decompose complex information needs, gather evidence from external sources, and maintain a reliable research state under multiple constraints. A central challenge is that early mistakes can be amplified over the search trajectory: agents may anchor on an incorrect candidate, collect partially supporting evidence, overlook contradictions, and accumulate errors that misguide subsequent reasoning. We propose \textbf{\ModelName{}}, a falsification-aware framework that formulates deep search as a propose--falsify process. \ModelName{} adopts a collaborative Searcher--Falsifier architecture, where the Searcher proposes candidate answers or sub-answers, while independent Falsifiers actively seek counterevidence, check unmet constraints, identify conflicts, and produce concise falsification reports for research-state revision. To strengthen this loop, we introduce falsification-aware training, combining SFT warmup on falsification trajectories with agentic RL based on parallel rollouts and hierarchical credit assignment over research stages, state transitions, and actions. Experiments on five search benchmarks show that \ModelName{} improves long-horizon deep search performance over backbone and harness ablations, while reducing confirmation bias and error accumulation through more effective falsification and research-state revision.
\end{abstract}

\section{Introduction}

Deep search requires large language model (LLM) agents to decompose complex, multi-constraint queries, retrieve and synthesize evidence from heterogeneous and potentially conflicting sources~\citep{xia2025open,zheng2025deepresearcher,press2023measuring}.
Unlike single-step retrieval-augmented generation, deep search requires agents to continuously refine candidate hypotheses over long-horizon interactions, verify task constraints, resolve conflicting evidence, and update the evolving research context~\citep{lewis2020retrieval,he2025webseer}.
Therefore, long-horizon deep search demands not only stronger retrieval capability but also the ability to maintain a reliable and consistent research state throughout an iterative search process, making it a key capability for advancing LLMs from conversational assistants toward autonomous knowledge-exploration agents~\citep{yao2026arc}.

However, the difficulty of long-horizon deep search lies not merely in executing more search steps.
Since each step depends on and modifies the research state, early mistakes can be amplified as the trajectory unfolds~\citep{xia2026diagnosing}.
We identify two sequential, mutually reinforcing failure modes underlying this process.
First, \textbf{confirmation bias} leads the agent to repeatedly seek evidence for an early candidate while rarely attempting to falsify it or test its underlying constraints, allowing incorrect hypotheses to persist.
These incorrect candidates, weakly supported claims, and unresolved contradictions are then incorporated into the research context, causing \textbf{error accumulation} that misguides subsequent search and reasoning.
The corrupted research state further reinforces the agent's search bias, forming a self-reinforcing error loop.
Thus, the challenge for deep search agents is not simply to search deeper, but to break this error loop by maintaining, questioning, and revising a reliable research state.

\begin{figure}[h]
\centering
\includegraphics[width=0.95\columnwidth]{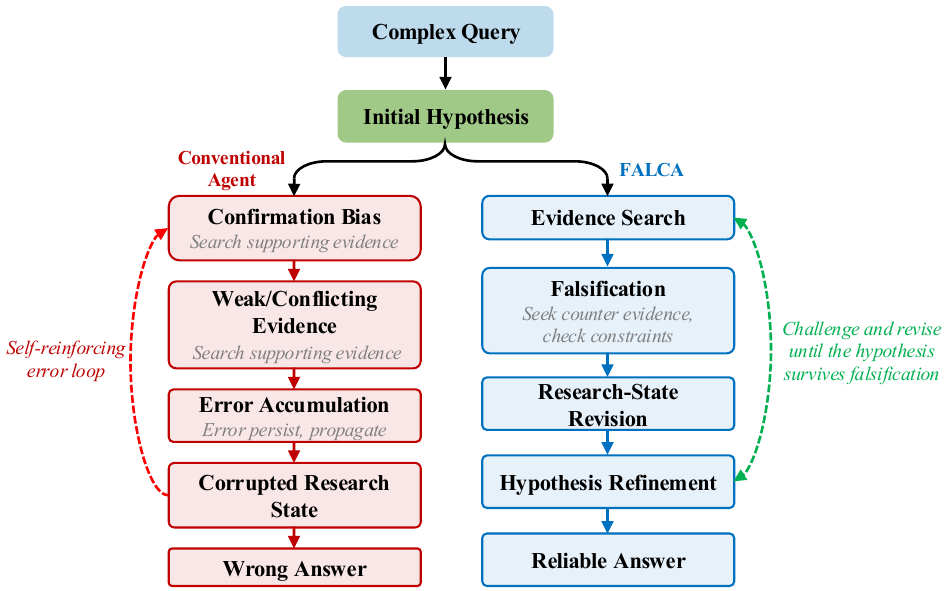}
\caption{\textbf{Motivation for \ModelName{}}. Conventional agents may fall into a self-reinforcing error loop, where confirmation bias leads to error accumulation and corrupts the research state. \ModelName{} breaks this loop through iterative falsification, research-state revision, and hypothesis refinement.}
\label{fig:motivation}
\end{figure}

Existing agents are typically optimized through trajectory imitation or final-answer rewards to search deeper along promising directions, but are rarely trained to challenge hypotheses, seek counterevidence, resolve conflicts, or revise the research state.
This is problematic because long-horizon deep search resembles a constrained research process rather than a linear retrieve-and-answer pipeline. The agent must propose candidates, test them against heterogeneous evidence, eliminate those that violate task constraints, and revise intermediate conclusions when conflicts emerge. Reliable progress therefore requires not only collecting supporting evidence but also identifying evidence that would invalidate current hypotheses, echoing the role of falsification in scientific inquiry~\citep{popper1963conjectures}. We thus formulate long-horizon deep search as a \textbf{propose--falsify} process, in which the agent periodically alternates among \emph{hypothesis candidate generation}, \emph{active falsification}, and \emph{research-state revision}. Under this view, falsification is not post-hoc verification but a context-management mechanism for checking whether the current research direction and candidate hypotheses remain valid before the agent searches deeper.

In this work, we propose \textbf{\ModelName{}}, a falsification-based agentic framework for long-horizon deep search. \ModelName{} adopts a collaborative Searcher--Falsifier architecture. The Searcher decomposes the problem, gathers evidence from external sources, maintains the research context, and proposes candidate answers. At periodic checkpoints, these candidates are passed to multiple Falsifier agents, each operating in an independent context to adversarially stress-test the candidate by seeking counterevidence, checking unmet constraints, identifying hidden conflicts, and suggesting revised search directions. Each Falsifier returns a concise falsification report summarizing the candidate's reliability, relevant evidence, unresolved or violated constraints, and recommended next steps. These reports are accumulated as a compact historical research state and fed back to the Searcher, which refreshes its working context and continues the propose--falsify loop until the information need is resolved.

To strengthen this propose--falsify research loop, we adopt a two-stage paradigm consisting of SFT warmup followed by falsification-aware agentic RL.
The model is first trained on falsification-aware trajectories to acquire basic capabilities for search, verification, counterevidence seeking, and context revision.
We then perform parallel rollouts for each problem to construct diverse trajectories that include successful, unsuccessful, and behaviorally distinct falsification paths.
Through hierarchical reward modeling and credit assignment, trajectory-level task feedback is further decomposed across research stages, state transitions, and individual actions.
Compared with agentic RL methods that rely primarily on final outcome rewards, \ModelName{} can reinforce search, falsification, and conflict-resolution behaviors that genuinely improve the research state, while avoiding indiscriminate reinforcement of redundant steps in successful trajectories or uniform penalization of locally valuable actions in failed trajectories.
Finally, we optimize the policy using a PPO-style clipped objective with shaped token-level advantages and KL regularization to the SFT reference policy.

Our contributions are threefold:
\begin{itemize}
\item We formulate long-horizon deep search as a falsification-centered propose--falsify process and instantiate it with a collaborative Searcher--Falsifier architecture, which periodically challenges candidate hypotheses and revises the research state to mitigate confirmation bias and error accumulation.

\item We develop a falsification-aware agentic RL framework that combines SFT warmup, parallel rollouts, and hierarchical credit assignment over trajectories, research stages, and actions, enabling the policy to learn search, falsification, and state-revision behaviors that improve the evolving research state.

\item We conduct extensive experiments on five search benchmarks, showing that \ModelName{} improves answer correctness and long-horizon search behavior, with ablations validating the contributions of falsification, research-state revision, and hierarchical credit assignment.

\end{itemize}

\section{Related Work}

\paragraph{Long-Horizon Search Agents.}
Retrieval-augmented generation grounds LLMs in external knowledge, while WebGPT, ReAct, and IRCoT extend this paradigm by interleaving reasoning with retrieval, browsing, and tool interaction~\citep{lewis2020retrieval,nakano2021webgpt,yao2023react,trivedi2023interleaving}.
More recent work trains LLMs to determine when and how to search through multi-turn interaction, using outcome-based reinforcement learning in controlled retrieval environments or end-to-end optimization over the open web~\citep{song2025r1searcher,jin2025searchr1,chen2025research,zheng2025deepresearcher}.
Complementary approaches study active context management for long-horizon information seeking~\citep{yao2026arc}.
These methods improve search depth and adaptability, but primarily optimize information acquisition or final-answer quality; the evolving research state is seldom modeled as a collection of falsifiable hypotheses whose supporting evidence, contradictions, and unmet constraints must be explicitly maintained and revised.

\paragraph{Reflection, Verification, and Falsification.}
Related work improves model reliability through iterative self-feedback, episodic reflection, tool-grounded critique, or learned reflection signals~\citep{madaan2023selfrefine,shinn2023reflexion,gou2024critic,asai2024selfrag}.
Web-agent methods incorporate reasoning behaviors such as reflection, lookahead, branching, and rollback~\citep{hu2025webcot}.
However, these mechanisms typically refine a generated output, recover from observed failures, or assess retrieved evidence, rather than treating falsification as a persistent search operation.
In contrast, our method explicitly assigns a Falsifier to seek disconfirming evidence, test unresolved constraints, expose contradictions, and trigger research-state revision before final synthesis, thereby addressing confirmation bias throughout the trajectory rather than only verifying an already formed answer.

\paragraph{Agentic Reinforcement Learning and Credit Assignment.}
Search-oriented agents are commonly initialized through supervised imitation and optimized with trajectory-level rewards.
Outcome-based RL can elicit multi-turn retrieval, planning, and self-correction~\citep{song2025r1searcher,jin2025searchr1,chen2025research,zheng2025deepresearcher}, but delayed rewards provide limited supervision for distinguishing critical verification and revision from redundant exploration.
Recent methods address this issue through hierarchical group-relative advantages, trajectory decomposition, and agent-specific process reward models~\citep{feng2025gigpo,luo2025agentlightning,xi2025agentprm}.
Unlike generic action-quality or goal-progress supervision, \ModelName{} assigns credit to research-state transitions, reinforcing actions that invalidate unsupported hypotheses, reconcile conflicting evidence, recover missing constraints, or improve research-state reliability.

\section{Methodology}

\begin{figure*}[h]
\centering
\includegraphics[width=0.9\linewidth]{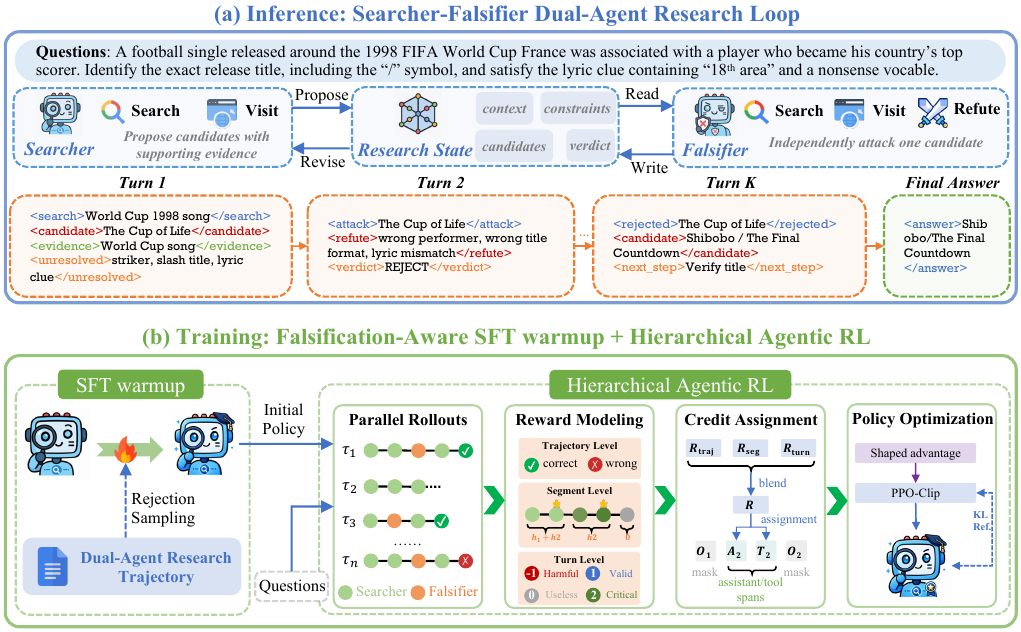}
\caption{\textbf{Overview of \ModelName{}}. The upper panel illustrates the Searcher-Falsifier dual-agent research loop for falsification-aware inference. The lower panel presents the training pipeline, consisting of SFT warmup, parallel rollouts, hierarchical reward modeling, credit assignment, and policy optimization.}
\label{fig:method}
\end{figure*}

\subsection{Problem Formulation}
We formulate long-horizon deep search as a sequential decision process.
Given a query $q$ and answer $y^*$, the agent interacts with an external environment to generate:
\begin{equation}
    \tau = (s_0, a_1, o_1, \cdots, a_T, o_T),
\end{equation}
where $s_t$, $a_t$, $o_t$ denote the research state, action, and observation.
The goal is to learn a policy $\pi_\theta(a_t|s_t)$ for search, verification, and answer synthesis under limited budgets.

We further represent long trajectories at multiple granularities.
A turn denotes one interaction step, while consecutive turns are grouped into segments:
\begin{equation}
    \tau = (g_0, g_1,\cdots,g_K).
\end{equation}
Segments serve as basic units for research state evolution, reward modeling, and credit assignment.
For multi-hop tasks, we model latent sub-goals: $q\to e_1\to e_2 = y^*$, where intermediate entity discovery provides process-level supervision.

\subsection{Tool Design}
To support long-horizon deep search, we equip the Searcher and Falsifier with a shared evidence environment.
Both agents access the same retrieval and browsing tools to acquire consistent evidence, while maintaining different objectives and action spaces.
The Searcher explores and maintains the research state, while the Falsifier seeks contradictions and returns structured feedback.
Specifically, \ModelName{} contains five tools:
\begin{itemize}
    \item \textbf{Search}: Retrieves relevant documents from external sources using batched queries for evidence discovery.
    \item \textbf{Visit}: Summarizes retrieved webpages into goal-conditioned evidence.
    \item \textbf{Update Context}: Compresses long interaction histories into structured research states containing verified facts, unresolved constraints, and future plans.
    \item \textbf{Finish}: Terminates the Searcher trajectory by producing the final answer with supporting evidence.
    \item \textbf{Refute}: Enables the Falsifier to return structured adversarial reports, including rejected constraints, evidence, and recommended follow-up searches.
\end{itemize}

\subsection{Falsification-Guided Research Trajectory Construction}
Generative deep-search agents may prematurely accept plausible but weakly supported candidates and repeatedly propagate them into subsequent contexts, leading to confirmation bias and error accumulation. To mitigate this issue, we construct falsification-guided research trajectories in which candidate hypotheses proposed by the Searcher are actively challenged by the Falsifier before being incorporated into later research states.

At the $k$-th research stage, the Searcher generates a candidate set from the current research state $S_k$: $\mathcal{H}_k = \{h_k^1,\cdots, h_k^{M_k}\}$.
For each candidate $h_k^m$, an independent Falsifier follows a contradiction-first strategy to search for counterevidence and returns a structured report through the \texttt{refute} tool: $F_k^m = (v_k^m, C_\text{ref}^m, C_\text{unr}^m, E_k^m, Q_\text{next}^m)$, where $v_k^m\in\{\texttt{reject}, \texttt{survive}, \texttt{uncertain}\}$ denotes the falsification verdict, $C_\text{ref}^m$ and $C_\text{unr}^m$ are the refuted and unresolved constraints, $E_k^m$ denotes the retrieved evidence, and $Q_\text{next}^m$ contains recommended follow-up queries.
The falsification reports are then written back to the research state: $S_{k+1} = \mathcal{U}\left(S_k, \mathcal{H}_k, \{F_k^m\}_{m=1}^{M_k}\right)$.

Rejected candidates are removed, unresolved constraints are converted into new search targets, and supporting or contradicting evidence is preserved through \texttt{update\_context}. Each context update defines a segment boundary, resulting in a trajectory composed of iterative exploration, falsification, and revision stages until the Searcher invokes \texttt{finish}.

Before agentic RL, we generate trajectories under the above protocol and apply rejection sampling to retain trajectories with correct answers, valid tool interactions, and sufficient evidence support.
The retained trajectories are used for SFT warmup with the masked language-modeling objective:
\begin{equation}
    \mathcal{L}_\text{SFT} = -\sum_i\sum_j m_{i,j}\log\pi_\theta(x_{i,j}|x_{i,<j}),
\end{equation}
where $m_{i,j}$ masks environmental observations such as search results and webpage contents.
This stage initializes the policy with basic capabilities in tool use, context updating, Searcher-Falsifier coordination, and candidate revision, providing a stable starting point for falsification-aware agentic RL.

\subsection{Falsification-Aware Agentic Reinforcement Learning}

\subsubsection{Parallel Falsification-Aware Rollout}
At each RL iteration, given a query $q$, we sample $n$ research trajectories in parallel from the current policy $\pi_\theta$: $\mathcal{G}(q) = \{\tau_i\}_{i=1}^n$.
Each trajectory executes the complete Searcher–Falsifier interaction loop online, with context compression enabled throughout the rollout.
Parallel sampling explores diverse candidate hypotheses, evidence paths, and falsification outcomes, providing varied samples for subsequent policy optimization.

To improve the stability and efficiency of long-horizon online rollouts, we employ two asynchronous mechanisms.
First, \textbf{asynchronous trajectory execution} allows different trajectories and their tool interactions to progress independently under trajectory-level time limits, preventing slow webpage visits or lengthy falsification rounds from blocking the entire rollout batch.
Second, \textbf{asynchronous process evaluation} computes turn-level process scores once a trajectory is completed while the remaining trajectories continue generation. This design provides denser process supervision without substantially extending the critical rollout path.

Finally, each completed trajectory is divided into multiple segments at the boundaries defined by \texttt{update\_context}. These parallel trajectories and their segment-level structures serve as the training units for the hierarchical reward modeling and fine-grained credit assignment introduced below.

\subsubsection{Hierarchical Reward Modeling and Credit Assignment}
Sparse outcome rewards provide limited process supervision for long-horizon deep search~\citep{qi2025webrl}.
A research trajectory may contain numerous search, verification, and state-update steps, while binary final-answer feedback cannot distinguish critical actions from redundant exploration within successful trajectories, nor identify locally valuable search behaviors in failed trajectories.
We therefore model rewards at progressively finer granularities, from trajectory outcomes and segment-level research progress to local turn-level action quality, and assign these signals to policy-generated tokens.

The trajectory-level reward measures final task success. Given the predicted answer $\hat{y}$ and the ground-truth answer $y^*$, we define $R_\text{traj} = 1[\hat{y}\equiv y^*]$, where answer correctness is jointly determined by exact matching and an LLM judge.
This reward provides the global task objective but does not distinguish the contributions of individual research steps.

We partition a complete trajectory into $K$ research segments using \texttt{update\_context}.
For a multi-hop task, let $h_j$ denote the segment in which the $j$-th subgoal is first completed, with $h_0=0$.
The reward weight $w_j$ associated with the $j$-th hop is uniformly distributed over the research stage between the previous and current hops:
\begin{equation}
    R_\text{seg}^k = \sum_{j=1}^J\frac{w_j}{h_j-h_{j-1}}1[h_{j-1}<k\leq h_j].
\end{equation}
This design provides denser credit to the consecutive research stages that contribute to subgoal completion, while avoiding the propagation of later-hop rewards to unrelated earlier behaviors. Segments after the final hop receive no segment-level reward.
To further assess local action quality, we use an LLM judge to score each turn: $s_t\in\{-1, 0, 1, 2\}$, corresponding to harmful, neutral, useful, and critical actions, respectively. The score is normalized as
\begin{equation}
    R_\text{turn}^t = \frac{\text{clip}(s_t, -1, 2) + 1}{3}.
\end{equation}
The evaluation considers not only retrieval relevance, but also whether an action acquires reliable evidence, identifies unmet constraints, falsifies an unsupported hypothesis, resolves evidence conflicts, or induces a meaningful research-state revision.
Finally, we assign the three reward signals to their corresponding policy tokens.
For a token $x_i$ belonging to segment $k(i)$ and turn $t(i)$, we define its shaped advantage as
\begin{equation}
    A_i^\text{shape} = \lambda_\text{traj}\frac{R_\text{traj}}{\sqrt{K}} + \lambda_\text{seg}R_\text{seg}^{k(i)} + \lambda_\text{turn}R_\text{turn}^{t(i)},
\end{equation}
where $\lambda_\text{traj} + \lambda_\text{seg} + \lambda_\text{turn} = 1$.

The trajectory-level reward is broadcast across the entire trajectory and normalized by $\sqrt{K}$ to prevent trajectories with more segments from receiving disproportionately large global credit. The segment-level reward is assigned to policy tokens within the corresponding research stage, while the turn-level reward is applied only to the associated turn span.
Environmental observations, including search results and webpage contents, are fully masked. The resulting token-level shaped advantages are used for policy optimization in the following section.

\subsubsection{Policy Optimization}
We optimize the policy using the token-level shaped advantages, without applying group-wise relative normalization.
This choice is motivated by the sparse and nearly binary outcomes of long-horizon deep search.
For difficult queries, trajectories sampled from the same prompt may receive identical final rewards, causing group-relative advantages to vanish.
In contrast, the shaped advantages preserve learning signals from segment-level progress and turn-level action quality, even when all trajectories within a rollout group have the same outcome.

For token $x_{i,t}$ in the $i$-th trajectory, we define the importance ratio as
\begin{equation}
    r_{i,t}(\theta) = \frac{\pi_\theta(x_{i,t}|x_{i,<t})}{\pi_{\theta_\text{old}}(x_{i,t}|x_{i,<t})}.
\end{equation}
We then optimize the policy using a PPO-style clipped objective~\citep{schulman2017proximal}:
\begin{equation}
\bar{r}_{i,t}(\theta)
=
\operatorname{clip}\!\left(
r_{i,t}(\theta),
1-\epsilon_{\mathrm{low}},
1+\epsilon_{\mathrm{high}}
\right).
\end{equation}

\begin{equation}
\mathcal{L}_{\mathrm{policy}}(\theta)
=
-\mathbb{E}_{i,t}
\left[
\min\left(
r_{i,t}(\theta)A_{i,t}^{\mathrm{shape}},
\bar{r}_{i,t}(\theta)A_{i,t}^{\mathrm{shape}}
\right)
\right].
\end{equation}
Following the asymmetric clipping strategy used in DAPO~\citep{yu2026dapo}, we adopt a slightly larger upper clipping bound, allowing the policy to more sufficiently increase the probabilities of high-value search, falsification, and research-state revision actions. To constrain policy drift from the supervised initialization, we further introduce token-level KL regularization with respect to a fixed SFT reference policy $\pi_\text{ref}$:
\begin{equation}
    \mathcal{L}(\theta) = \mathcal{L}_\text{policy}(\theta) + \beta\mathbb{E}_{i,t}[D_\text{KL}(\pi_\theta(\cdot|x_{i,<t})\|\pi_\text{ref}(\cdot|x_{i, <t}))].
\end{equation}
This optimization preserves the exploration benefits of parallel rollouts while avoiding the loss of learning signals caused by degenerate rollout groups. The clipped objective and reference-policy regularization further stabilize policy updates during long-horizon agentic reinforcement learning.

\begin{table*}[t]
    \centering
    \small
    \renewcommand{\arraystretch}{1.10}
    \setlength{\tabcolsep}{4.5pt}

    \begin{tabular*}{\textwidth}{
        @{\extracolsep{\fill}}lccccc@{}
    }
        \toprule
        \textbf{Model}
        & \textbf{BrowseComp}
        & \textbf{GAIA}
        & \textbf{DeepSearchQA}
        & \textbf{xbench-2510}
        & \textbf{WideSearch-en} \\
        \midrule

        \rowcolor{gray!12}
        \multicolumn{6}{@{}l@{}}{
            \textbf{Default Context-Management Harness}
        } \\

        GPT-5
        & 54.9
        & 76.4
        & 79.0
        & 75.0
        & 59.7 \\

        Opus-4.6
        & 83.7
        & 75.0
        & 91.3
        & --
        & 77.5 \\

        Gemini-3.1-Pro
        & 85.9
        & 80.6
        & 93.3
        & 53.0
        & 66.4 \\

        Qwen3.5-397B
        & 78.6
        & 83.5
        & 82.1
        & 61.0
        & 74.0 \\

        DeepMiner-32B
        & 33.5
        & 58.7
        & --
        & 62.0
        & -- \\

        WebDancer-32B
        & 3.8
        & 51.5
        & --
        & 39.0
        & -- \\

        WebSailor-32B
        & 10.5
        & 53.2
        & --
        & --
        & -- \\

        Tongyi-DeepResearch-30B
        & 43.4
        & 70.9
        & --
        & 55.0
        & -- \\

        Qwen3.5-9B
        & 34.2
        & 65.0
        & 65.5
        & 51.0
        & 17.5 \\

        MiroThinker-8B
        & 31.1
        & 66.4
        & 36.7
        & 34.0
        & -- \\

        WebExplorer-8B
        & 15.7
        & 32.0
        & 17.8
        & 53.7
        & -- \\

        LiteResearcher-4B
        & 27.5
        & 71.3
        & 17.2
        & 12.0
        & 1.73 \\

        AgentCPM-Explore-4B
        & 24.1
        & 63.9
        & 32.8
        & 34.0
        & -- \\

        Qwen3.5-4B
        & 32.7
        & 65.0
        & 62.6
        & 45.0
        & 53.9 \\

        \midrule

        \rowcolor{gray!12}
        \multicolumn{6}{@{}l@{}}{
            \textbf{Falsifier-Based Harness}
        } \\

        Qwen3.5-9B
        & 41.4
        & 68.0
        & 65.0
        & 48.0
        & \textbf{67.2} \\

        Qwen3.5-4B
        & 39.0
        & 62.1
        & 63.3
        & 49.0
        & 64.3 \\

        \textbf{FALCA-4B}
        & \textbf{56.0}
        & \textbf{72.8}
        & \textbf{70.8}
        & \textbf{58.0}
        & 65.8 \\

        \bottomrule
    \end{tabular*}

    \vspace{2pt}
    \begin{minipage}{\textwidth}
        \scriptsize
        \textit{Notes.}
        Best open-source results are bolded.
        A dash denotes an unavailable or pending result.
    \end{minipage}

    \caption{Main experimental results on five search benchmarks.}
    \label{tab:main-results}
\end{table*}

\section{Experiments}
\subsection{Experimental Setup}
\subsubsection{Datasets and Baselines}
We evaluate \ModelName{} on five search-augmented reasoning benchmarks: \textbf{BrowseComp} and \textbf{DeepSearchQA} for long-horizon retrieval and evidence synthesis~\cite{wei2025browsecomp,gupta2026deepsearchqa}; \textbf{GAIA} and \textbf{xbench-2510} for agentic planning, tool use, and multi-step reasoning~\cite{mialon2024gaia,chen2025xbench}; and \textbf{WideSearch-en} for broad-coverage retrieval and synthesis~\cite{wong2025widesearch}.
We report Item-F1 for WideSearch-en, F1 for DeepSearchQA, and accuracy for the remaining benchmarks, with all scores reported in percentage points.

We compare with both closed-source and open-source baselines, including GPT-5~\cite{singh2025openai}, Opus-4.6~\cite{anthropic2026opus46}, Gemini-3.1-Pro~\cite{google2026gemini31}, Qwen3.5-397B~\cite{qwen2026qwen35}, and recent search-oriented agents such as DeepMiner~\cite{tang2025beyond}, WebDancer~\cite{wu2026webdancer}, WebSailor~\cite{li2025websailor}, Tongyi-DeepResearch~\cite{team2025tongyi}, MiroThinker~\cite{team2025mirothinker}, WebExplorer~\cite{liu2025webexplorer}, LiteResearcher~\cite{li2026literesearcher}, and AgentCPM-Explore~\cite{chen2026agentcpm}. We also report Qwen3.5-4B and Qwen3.5-9B as backbone baselines.

\subsubsection{Implementation Details}
We initialize the Searcher from an SFT-trained Qwen3.5-4B and optimize it with reinforcement learning, while the Falsifier uses the same backbone in an independent context.
Turn-level process evaluation, final-answer judging, and webpage and context summarization are handled by Qwen3.5-35B-A3B.
The \texttt{search} tool retrieves top-ranked titles, URLs, and snippets from a local HNSW index built with BGE-m3 embeddings~\citep{chen2024bge,malkov2018efficient}.
Training and rollouts use eight NVIDIA H100 80GB GPUs.

\subsection{Main Results}

As shown in Table~\ref{tab:main-results}, introducing the
Falsifier-based harness yields clear overall gains while keeping the Qwen3.5-4B backbone unchanged.
Compared with the default context-management harness, its average score increases from \(51.8\) to \(55.5\).
This result suggests that, beyond collecting supporting evidence, periodically challenging candidates, checking unmet constraints, and injecting falsification reports into the research state can improve search reliability.
This observation is consistent with our motivation: active falsification helps mitigate confirmation bias around early hypotheses and reduce the accumulation of incorrect candidates and unresolved contradictions over long-horizon interactions.

Under the same Falsifier-based harness, \ModelName{}-4B improves the average score from \(55.5\) to \(64.7\), outperforming the falsification-untrained Qwen3.5-4B on all benchmarks.
This indicates that merely introducing a Falsifier is insufficient to fully realize the benefits of falsification.
SFT warmup teaches the Searcher the basic propose--falsify protocol, while agentic RL optimizes how it incorporates falsification reports, revises the research state, and redirects subsequent retrieval.
Notably, \ModelName{}-4B also surpasses Qwen3.5-9B under the same harness (\(64.7\) vs.\ \(57.9\)) and remains competitive with several 30B--32B search agents.
These results suggest that falsification-aware training can be more parameter-efficient than simply scaling up the backbone.

\ModelName{}-4B achieves its largest improvements on BrowseComp, xbench-2510, and DeepSearchQA, which require multi-step evidence aggregation, constraint verification, candidate elimination, and research-state revision.
This pattern suggests that falsification is beneficial for constraint-intensive deep-search tasks.
On WideSearch-en, which emphasizes broad-coverage information collection, the improvement is smaller, although \ModelName{}-4B still achieves a strong absolute score of \(65.8\).
Overall, \ModelName{}-4B provides the largest benefits on falsification-intensive tasks while preserving strong broad-coverage retrieval performance.

\subsection{Ablation Study}
\begin{figure}[h]
\centering
\includegraphics[width=0.9\columnwidth]{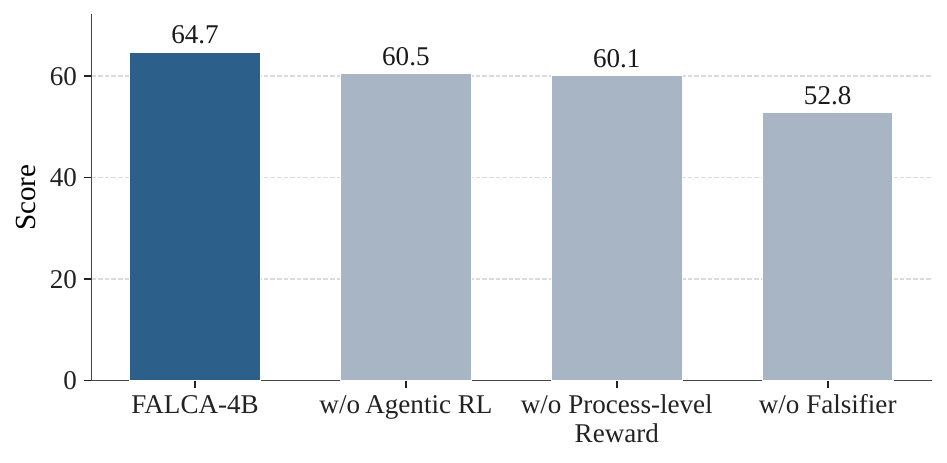}
\caption{Ablation study of the key inference and training components in \ModelName{}.}
\label{fig:ablation_study}
\end{figure}

To assess the key inference and training components in \ModelName{}, we construct three ablated variants.
\emph{w/o Falsifier} removes the Falsifier and its report injection, evaluating the role of active falsification and hypothesis revision.
\emph{w/o Agentic RL} retains only the SFT warmup, while \emph{w/o Process-level Reward} removes the segment- and turn-level rewards and relies solely on trajectory-level feedback.
All variants use the same backbone, training data, inference budget, and evaluation protocol.
As shown in Figure~\ref{fig:ablation_study}, \ModelName{} achieves the highest score of \(64.7\).
Removing the Falsifier causes the largest drop to \(52.8\), highlighting the importance of seeking counterevidence, identifying constraint violations, and revising incorrect hypotheses.
Without agentic RL, the score decreases to \(60.5\), indicating that SFT establishes the falsification protocol, while RL improves how the agent uses falsification feedback during multi-turn search.
Removing process-level rewards reduces the score to \(60.1\), suggesting that terminal outcomes provide insufficient credit for intermediate evidence acquisition, candidate elimination, and research-state revision.
Overall, the Falsifier provides the core inference mechanism, while agentic RL and process-level rewards enable it to be learned and executed more effectively.

\begin{figure*}[t]
\centering
\includegraphics[width=0.9\linewidth]{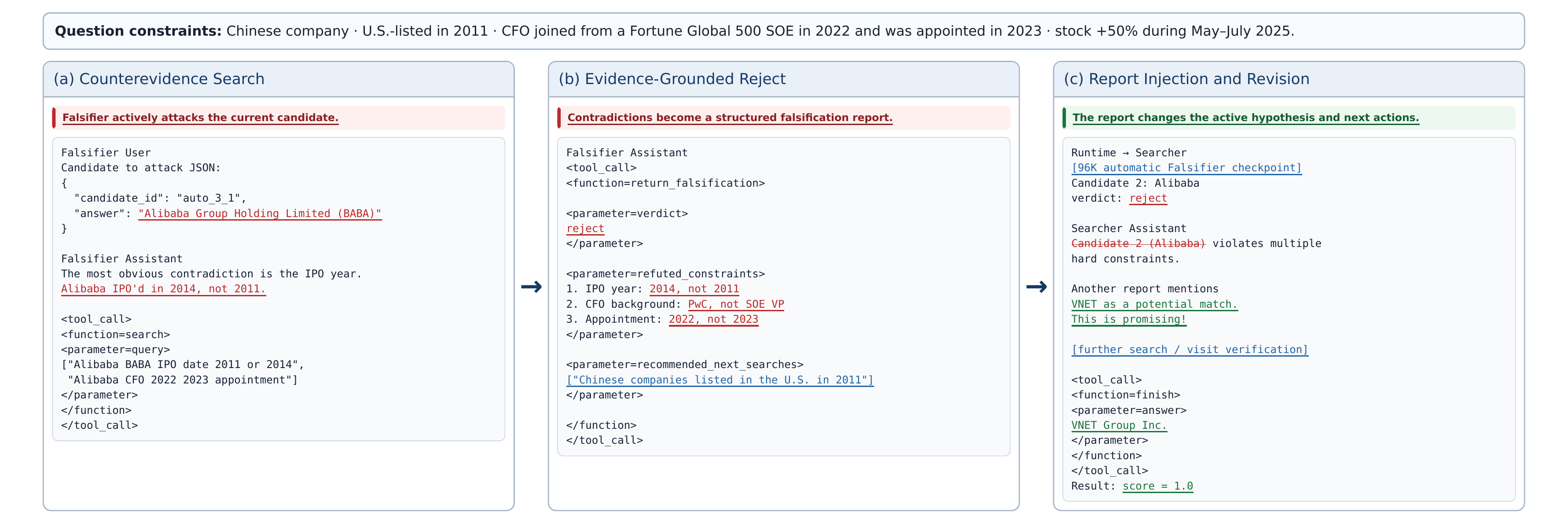}
\caption{\textbf{A real falsification-driven revision trajectory from xbench-2510.}
The Falsifier refutes Alibaba by identifying violations in its listing year, CFO background, and appointment date.
After the report is injected, the Searcher adopts VNET as its new hypothesis, verifies the remaining constraints, and returns the correct answer.}

\label{fig:falsification}
\end{figure*}

\subsection{Behavioral Analysis}

\subsubsection{Falsification-Driven Revision}
We examine whether Falsifier reports induce actionable research-state revisions rather than merely provide additional verification.
As shown in the real xbench-2510 trajectory in Figure~\ref{fig:falsification}, the injected report causes the Searcher to discard an incorrect hypothesis, revise its research state, and redirect retrieval toward the correct candidate.
This case provides qualitative evidence that falsification functions as an actionable state transition in \ModelName{}.

\subsubsection{Long-Horizon Test-Time Scaling}
\begin{figure}[h]
\centering
\includegraphics[width=0.9\columnwidth]{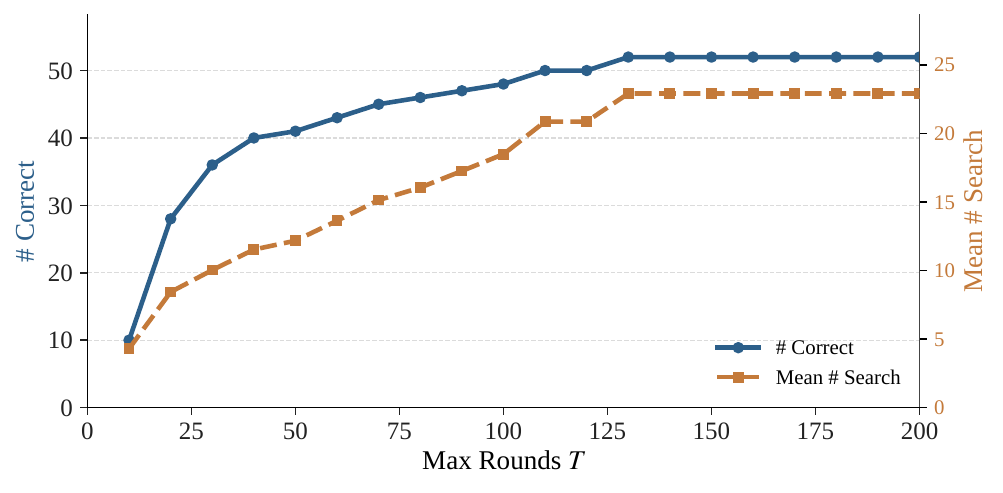}
\caption{Long-horizon test-time scaling on BrowseComp.}
\label{fig:tts}
\end{figure}

To study depth-based test-time scaling on BrowseComp, we fix the model and tool configuration and measure the cumulative number of correct trajectories and mean search calls at each round threshold \(T\).
As shown in Figure~\ref{fig:tts}, \(50\%\) of correct trajectories finish within 20 rounds and \(89\%\) within 120 rounds. Extending the threshold to 200 rounds increases coverage only to \(93\%\), while mean search calls rise from 4.3 to 22.9.
These results demonstrate positive reasoning-depth scaling, but with diminishing returns beyond approximately 120--150 rounds.

\subsection{Training Dynamics and Analysis}
Long-horizon deep search exhibits sparse terminal rewards.
Complex queries require multiple rounds of retrieval, constraint verification, and evidence synthesis, while trajectories may terminate due to timeouts, interaction limits, or length truncation.
Consequently, trajectories sampled for the same query often receive identical rewards; in our training, approximately 66\% of rollout groups have zero within-group reward variance, weakening GRPO-style group-relative advantages.

\begin{figure}[h]
\centering
\includegraphics[width=0.9\columnwidth]{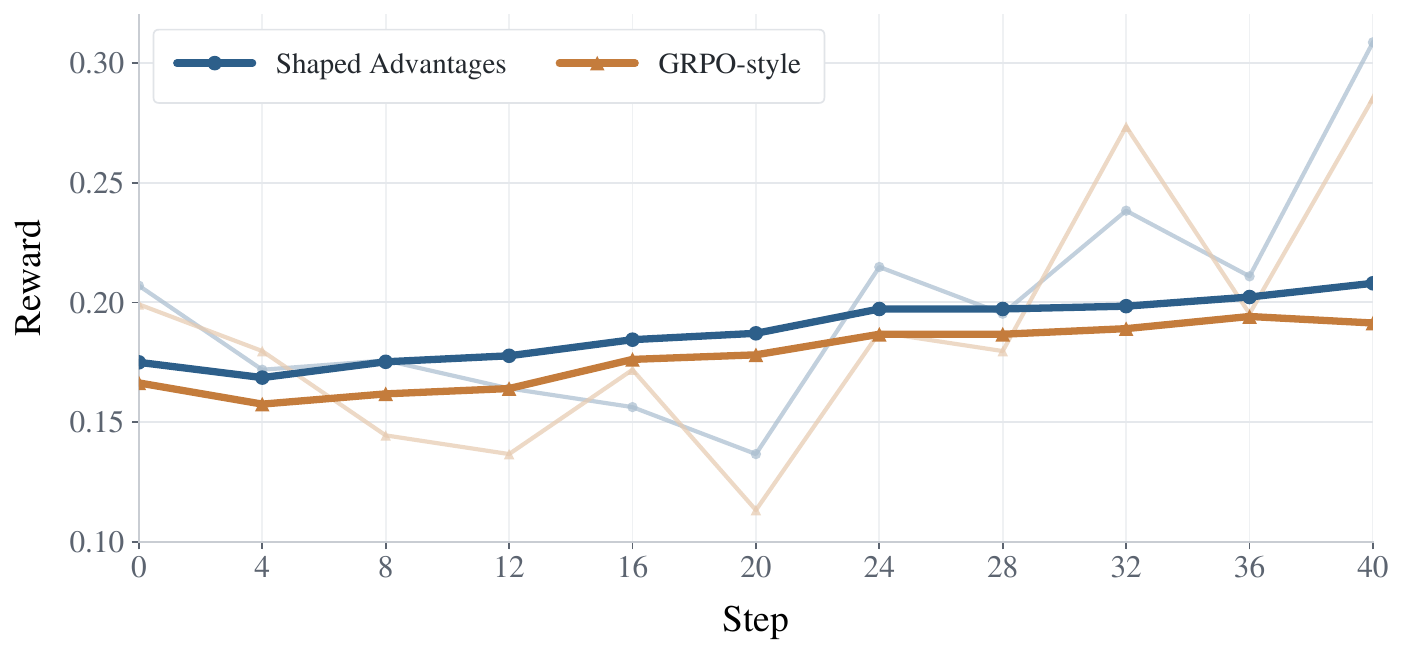}
\caption{Training reward dynamics of shaped and GRPO-style advantages under degenerate rollout groups.}
\label{fig:shaped_grpo_figure}
\end{figure}

In contrast, our shaped advantages combine trajectory-, segment-, and turn-level rewards to capture process-level differences within individual trajectories.
As shown in Figure~\ref{fig:shaped_grpo_figure}, the training reward of GRPO-style optimization reaches an early plateau, whereas shaped advantages exhibit a more sustained upward trend.
This suggests that fine-grained process signals can continue to provide useful supervision even when trajectories share the same terminal outcome.
As reported in Table~\ref{tab:shaped_grpo_table}, shaped advantages improve the average score from \(61.2\) to \(64.7\) and induce more turns, tokens, and search calls.
These results suggest that shaped advantages preserve informative learning signals under degenerate rollout groups, supporting more persistent long-horizon exploration.

\begin{table}[h]
    \centering
    \resizebox{\columnwidth}{!}{
    \begin{tabular}{lccccc}
    \toprule
    Setting & Avg. & Turns & Tokens & Search & Visit \\
    \midrule
    Shaped Advantages & \textbf{64.7} & 79.8 & 8.7M & 63.6 & 14.9 \\
    GRPO-style & 61.2 & 62.8 & 6.7M & 47.7 & 14.0 \\
    \bottomrule
    \end{tabular}
    }
    \caption{Comparison of task performance and search behavior between shaped and GRPO-style advantages.}
    \label{tab:shaped_grpo_table}
\end{table}

\section{Conclusion}
We introduced \ModelName{}, a falsification-aware reinforcement learning framework that formulates long-horizon deep search as an iterative propose--falsify process.
\ModelName{} combines independent Falsifiers for research-state revision with SFT warmup, parallel agentic rollouts and hierarchical credit assignment over trajectories, research segments, and individual actions.
Experiments on five benchmarks demonstrate consistent improvements over backbone and harness baselines with strong parameter efficiency.
Ablation studies identify the Falsifier as the primary source of performance gains and validate the contributions of agentic RL and process-level rewards.
Training and behavioral analyses show that shaped advantages preserve learning signals under degenerate rollout groups, while falsification reports can redirect the Searcher away from plausible but incorrect hypotheses.
Overall, our results suggest that reliable long-horizon deep search depends not only on searching deeper, but also on continually questioning and repairing the research state.

\bibliography{aaai2027}

\clearpage
\appendix

\section{Benchmark Details}
\subsection{Overall Coverage}
We evaluate \ModelName{} on five complementary benchmarks that cover diverse forms of search-augmented reasoning.
BrowseComp and DeepSearchQA emphasize long-horizon information seeking, requiring persistent retrieval, multi-source evidence aggregation, and answer synthesis.
GAIA evaluates general agentic capabilities, including planning, tool use, multimodal understanding, and multi-step reasoning, while xbench-2510 focuses more specifically on end-to-end deep search in the Chinese web environment.
WideSearch-en complements these depth-oriented benchmarks by evaluating broad information seeking, where an agent must collect and organize a large number of atomic facts from a wide search space.
Together, these benchmarks span constraint-intensive deep search, general tool-augmented reasoning, and broad-coverage information collection, enabling a comprehensive evaluation of search depth, constraint verification, research-state maintenance, and information coverage.

\subsection{BrowseComp}
BrowseComp consists of 1,266 open-domain questions designed to evaluate an agent's ability to persistently browse the web for hard-to-find and entangled information.
Although its reference answers are typically short and easily verifiable, they are difficult to obtain through direct lexical matching with the original question.
Solving these tasks often requires repeated query reformulation, integration of evidence distributed across webpages, and elimination of plausible candidates that satisfy only part of the given constraints.
BrowseComp therefore provides a focused evaluation of persistent retrieval, candidate verification, and constraint-intensive reasoning.

\subsection{GAIA}
GAIA is a general-assistant benchmark comprising 466 real-world questions with unambiguous answers. Its tasks require combinations of logical reasoning, web browsing, multimodal information processing, file handling, computation, and external tool use. Unlike benchmarks devoted exclusively to web retrieval, GAIA evaluates whether an agent can identify the operations required by a task, organize them into a coherent multi-step plan, and maintain task-relevant information across heterogeneous interactions.

\subsection{DeepSearchQA}
DeepSearchQA contains 900 handcrafted prompts across 17 domains and evaluates whether agents can execute complex, multi-step search plans to produce exhaustive answer sets. Each task is structured as a causal search chain in which later discoveries depend on the successful completion of earlier steps. In addition to retrieving fragmented information from multiple sources, agents must perform entity resolution and deduplication, maintain context over long trajectories, and determine when the open-ended search space has been sufficiently covered. We report F1, which balances the precision of the predicted answer set against its coverage of the reference answers.

\subsection{xbench-2510}
xbench-DeepSearch evaluates end-to-end search capabilities, including planning, retrieval, reasoning, and information summarization.
We use xbench-DeepSearch-2510, a refreshed 100-question version designed around the Chinese web environment.
Its questions are expert-curated and typically require agents to integrate multiple constraints, distinguish among partially matching candidates, and synthesize evidence into a uniquely verifiable answer.

\subsection{WideSearch-en}
WideSearch evaluates broad information seeking rather than deep investigation of a single hidden answer.
The full benchmark contains 200 manually curated tasks from more than 15 domains, equally divided between English and Chinese; we evaluate on its 100-question English subset, WideSearch-en.
Each task requires the agent to collect a large set of independently verifiable atomic facts and organize them into a structured output.
Its primary challenge is therefore to maximize information coverage while avoiding irrelevant, duplicated, or incorrect entries.
We report Item-F1, which jointly measures item-level precision and recall over the collected information.

\section{Baselines}
We compare \ModelName{} with representative search agents spanning model scales from 4B to 32B parameters.
These baselines cover diverse approaches to long-horizon information seeking, including challenging data synthesis, context management, interaction scaling, and agentic reinforcement learning.
This comparison evaluates both absolute search performance and parameter efficiency.
For external agents, we use results reported in their original papers or official technical reports only when the benchmark version and evaluation metric are compatible with ours; otherwise, the corresponding result is marked as unavailable.

\subsection{Large-Scale Agents}
We include four 30B--32B search agents.
\textbf{DeepMiner-32B} combines reverse-constructed, high-difficulty search tasks with a dynamic context-window mechanism to support sustained long-horizon interactions.
\textbf{WebDancer-32B} develops an end-to-end information-seeking agent through browsing-data construction, trajectory sampling, supervised fine-tuning, and reinforcement learning.
\textbf{WebSailor-32B} targets complex information-seeking tasks through high-uncertainty data construction, rejection-sampling fine-tuning, and agentic reinforcement learning.
\textbf{Tongyi-DeepResearch-30B} integrates agentic mid-training and post-training with scalable automatic data synthesis for long-horizon deep research.

\subsection{Compact Agents}
We further compare with four compact search agents.
\textbf{MiroThinker-8B} studies interaction scaling, training the model to benefit from deeper and more frequent agent--environment interactions.
\textbf{WebExplorer-8B} constructs challenging search tasks through model-based exploration and iterative long-to-short query evolution, followed by supervised fine-tuning and reinforcement learning.
\textbf{LiteResearcher-4B} introduces a lightweight virtual search environment to enable scalable and stable agentic RL for deep-research agents.
\textbf{AgentCPM-Explore-4B} targets long-horizon exploration at the edge-model scale through model fusion, reward denoising, and contextual information refinement.
The two 4B agents provide the most direct parameter-matched comparison with \ModelName{}-4B.

\section{Implementation Settings}

\section{Prompts}
The complete prompts used in our experiments are provided in
Figures~\ref{fig:searcher_system_prompt}--\ref{fig:answer_grading_prompt}.
We use role-specific prompts to separate hypothesis construction from
adversarial verification. The Searcher is instructed to maintain multiple
candidate hypotheses, distinguish hard constraints from soft clues, gather
URL-grounded evidence, and periodically externalize its research state
through \texttt{update\_context}. In contrast, the Falsifier follows a
contradiction-first protocol: it independently attacks one candidate,
checks whether the candidate violates the task constraints or requested
answer type, and returns a structured verdict of \texttt{reject},
\texttt{survive}, or \texttt{uncertain}.

We additionally use auxiliary prompts for evidence processing, context
management, and evaluation. The visit summarization prompt extracts
goal-conditioned evidence from webpages while producing a compact summary
for later context folding. When the context limit is reached, the context
limit prompt requires the Searcher either to serialize the current research
state or to terminate with a verified answer. Finally, the answer grading
prompt extracts the predicted answer and evaluates its correctness against
the reference answer under a standardized format.

\begin{figure*}
\begin{AIbox}{Searcher System Prompt}
\small
Search intensity is high.
You are the \textbf{main Proposer} in an adversarial Falsifier search loop.
Your task is to find the exact answer with evidence, not merely a plausible answer.

\textbf{Core policy}:
\begin{itemize}
[leftmargin=1.5em,itemsep=0.5pt,topsep=0.5pt,parsep=0.5pt]
    \item Maintain candidate answers when evidence permits. Track each candidate's support, unresolved constraints, and possible contradictions.
    \item Optimize for candidate recall before candidate precision. Keep several plausible candidates alive early, including weak or partially supported candidates, instead of collapsing to one anchor too soon.
    \item Explore multiple independent search frames when the task is ambiguous: exact-phrase retrieval, clue-to-document retrieval, entity/author intersection, source/domain-specific search, and broad synonym/paraphrase search. If one frame stalls, switch frames.
    \item Track assumptions you added that are not stated in the question, such as field, nationality, gender, time period, source type, or answer type. Actively search against these assumptions before trusting them.
    \item Separate hard constraints from soft clues. A hard contradiction should eliminate or sharply downgrade a candidate.
    \item Use discriminative searches that can refute candidates, separate close candidates, or verify unresolved hard constraints.
    \item Use \texttt{visit} for decisive evidence; snippets alone are not enough for final claims or candidate rejection.
    \item Use \texttt{update\_context} as a checkpoint: pass dense ``context'', ``constraints'', and a non-empty ``candidates'' list whenever any plausible candidate or answer frame exists. The runtime will run independent Falsifier agents and append their attack reports to the refreshed context.
    \item Avoid ``candidates: []'' unless you have explicitly tried at least two independent recall routes and still have no plausible candidate. If unsure, include weak candidates with clear unresolved constraints.
    \item Falsifier reports are strong evidence to integrate. Treat ``reject'' as a serious blocker unless later direct evidence overturns it; treat ``uncertain'' as a warning that more verification or candidate recall is needed; treat ``survive'' as useful only when hard constraints are directly supported.
    \item Do not finish with a candidate rejected or left uncertain by a Falsifier unless later evidence directly resolves the issue.
    \item Never simulate tool outputs. Always call the actual tools.
\end{itemize}
\textbf{Tools}:

... \{tool\_des\} ...
\end{AIbox}

\caption{System prompt for the Searcher agent.}
\label{fig:searcher_system_prompt}
\end{figure*}

\begin{figure*}
\begin{AIbox}{Searcher User Prompt}
\small
Question: \{question\}

Start with a search tool call.
Build evidence, constraints, and candidates as you go.

\textbf{Before finishing}:
\begin{itemize}
[leftmargin=1.5em,itemsep=0.5pt,topsep=0.5pt,parsep=0.5pt]
    \item every critical claim must have a cited URL;
    \item unresolved contradictions must be resolved or explicitly judged non-decisive;
    \item candidate answers should have survived direct falsification or later evidence must explain why a rejection/uncertainty was wrong;
    \item the final answer must match the answer type and field asked by the question, not a nearby affiliation, source, person, or related entity.
\end{itemize}
When context is long, search stalls, or candidate analysis reaches a milestone, call \texttt{update\_context} with ``context'', ``constraints'', and ``candidates''.
Prefer passing 2-5 plausible candidates or explicit answer frames; use ``[]'' only after explicit recall attempts fail.

When the answer is sufficiently verified, call \texttt{finish} with a concise ``answer'', JSON-array ``evidences'', and ``confidence''.

Do not answer from memory and do not write text after a tool call.
\end{AIbox}
\caption{User prompt for the Searcher agent.}
\label{fig:searcher_user_prompt}

\end{figure*}

\begin{figure*}
\begin{AIbox}{Falsifier System Prompt}
\small
You are an \textbf{adversarial Falsifier agent} for Deep Search.

Your primary task is to attack one candidate hypothesis proposed by the main agent.
Search for evidence that the candidate violates the question's hard constraints, answer type, source field, or implied entity role.
Do not choose the final answer and do not help the candidate first.

Also attack the main agent's framing when relevant. If the candidate seems to come from a wrong assumption, wrong answer field, nearby entity drift, or too-narrow search space, report that clearly and recommend alternative recall searches.

\textbf{Tools}:

... \{tool\_des\} ...

\textbf{Rules}:
\begin{itemize}
[leftmargin=1.5em,itemsep=0.5pt,topsep=0.5pt,parsep=0.5pt]
    \item Use the actual \texttt{search}/\texttt{visit} tools. Never invent tool outputs.
    \item Search for refuting evidence before supporting evidence.
    \item Visit authoritative pages for decisive claims; snippets alone are not enough for rejection.
    \item Distinguish hard contradiction from ambiguity or missing evidence.
    \item Check the original question's requested answer type carefully. A related person, institution, article, affiliation, event, or source is a refutation if the question asks for a different field.
    \item ``reject'' when a hard constraint is contradicted, the answer type/field is wrong, or a better-supported interpretation makes the candidate untenable.
    \item ``survive'' only when the candidate has direct support for the key hard constraints. Do not use ``survive'' merely because no contradiction was found.
    \item Use ``uncertain'' when evidence is incomplete, indirect, snippet-only, or when the candidate might be right but key constraints remain unverified.
    \item For no-answer or ``not found'' candidates, aggressively search for one counterexample. Missing search results are not enough for ``survive''.
    \item Return exactly one \texttt{Refute} call when done.
    \item The report fields are: candidate\_id, candidate\_answer, verdict, refuted\_constraints, supporting\_constraints, unresolved\_constraints, refuting\_sources, supporting\_sources, recommended\_next\_searches, summary.
    \item Allowed verdict values: reject, survive, uncertain.
    \item Do not output text after a tool call.
\end{itemize}

\end{AIbox}
\caption{System prompt for the Falsifier agent.}
\label{fig:falsifier_system_prompt}
\end{figure*}

\begin{figure*}
\begin{AIbox}{Falsifier User Prompt}
\small
\textbf{Original question}:\\
\texttt{\{question\}}

\textbf{Compressed main-agent context}:\\
\texttt{\{context\}}

\textbf{Constraint list JSON}:\\
\texttt{\{constraints\_json\}}

\textbf{Candidate to attack JSON}:\\
\texttt{\{candidate\_json\}}

Attack this single candidate.
Look for direct evidence that violates the hard constraints, answer type, source field, or makes another interpretation more likely.
If you find a decisive contradiction or wrong-field answer, return \texttt{reject}.
If contradiction search fails and key constraints are directly supported, return \texttt{survive}.
If evidence is incomplete, indirect, or only shows that searches failed, return \texttt{uncertain}.
Put alternative candidates, recall routes, and discriminative next searches in \texttt{recommended\_next\_searches}.
\end{AIbox}
\caption{User prompt for candidate falsification.}
\label{fig:falsifier_user_prompt}
\end{figure*}

\begin{figure*}
\begin{AIbox}{Visit Summarization Prompt}
\small
Please process the following webpage content and user goal to extract relevant information:

\textbf{Webpage Content}: \texttt{\{webpage\_content\}}

\textbf{User Goal}: \texttt{\{goal\}}

\textbf{Task Guidelines}:
\begin{itemize}[leftmargin=1.5em,itemsep=0.5pt,topsep=0.5pt,parsep=0.5pt]
    \item \textbf{Content Scanning for Rationale}: Locate the specific sections/data directly related to the user's goal within the webpage content.
    \item \textbf{Key Extraction for Evidence}: Identify and extract the most relevant information; never miss important information; output the full original context as far as possible; prefer a longer \texttt{evidence} field when the page contains multiple relevant details.
    \item \textbf{Summary Output for Summary}: Organize into a detailed summary with logical flow, preserving names, dates, institutions, locations, titles, and relationships whenever they help the goal.
    \item \textbf{Compact Summary for Folding}: Write a \texttt{compact\_summary} of 120--180 words for later context folding. It must be dense, structured, and preserve decisive entities, dates, locations, relationships, evidence polarity, and the page's contribution to the user goal.
    \item If the page contains several independently useful facts, include all of them instead of compressing too aggressively.
\end{itemize}

\textbf{Final output} (JSON fields): \texttt{rational}, \texttt{evidence}, \texttt{summary}, \texttt{compact\_summary}.
\begin{itemize}[leftmargin=1.5em,itemsep=0.5pt,topsep=0.5pt,parsep=0.5pt]
    \item \texttt{evidence}: can be long; preserve original wording/context as much as possible.
    \item \texttt{summary}: can be multiple paragraphs; should not be artificially shortened.
    \item \texttt{compact\_summary}: 120--180 words, formatted for reuse after context folding.
\end{itemize}
\end{AIbox}
\caption{Prompt for goal-conditioned webpage summarization.}
\label{fig:visit_summary_prompt}
\end{figure*}

\begin{figure*}
\begin{AIbox}{Context Limit Prompt}
\small
The context limit has been exceeded for the task.
You \textbf{MUST} either update the context now or finish now.
If the answer is not fully verified, call \texttt{update\_context} with the best current constraints and candidates so Falsifier agents can attack them.
Prefer weak candidates or answer frames over \texttt{candidates: []}.
If a Falsifier rejected or questioned the leading candidate, finish only if later direct evidence resolves that objection.

\textbf{Output requirements (STRICT)}:
\begin{itemize}[leftmargin=1.5em,itemsep=0.5pt,topsep=0.5pt,parsep=0.5pt]
    \item Output exactly one tool call.
    \item The only allowed tool call is \texttt{update\_context} or \texttt{finish}.
\end{itemize}
\end{AIbox}
\caption{Prompt for handling context-limit events.}
\label{fig:context_limit_prompt}
\end{figure*}

\begin{figure*}
\begin{AIbox}{Answer Grading Prompt}
\small
Judge whether the following \texttt{[response]} to \texttt{[question]} is correct or not based on the precise and unambiguous \texttt{[correct\_answer]} below.

\texttt{[question]}: \texttt{\{question\}}

\texttt{[response]}: \texttt{\{response\}}

Your judgement must be in the format and criteria specified below:

\textbf{extracted\_final\_answer}:
The final exact answer extracted from the \texttt{[response]}.
Put \texttt{None} if there is no exact final answer to extract.

\texttt{[correct\_answer]}: \texttt{\{correct\_answer\}}

\textbf{reasoning}:
Explain why the extracted final answer is correct or incorrect based on \texttt{[correct\_answer]},
focusing only on meaningful differences.
Do not comment on background, do not solve the problem, and do not argue for any answer different from \texttt{[correct\_answer]}.

\textbf{correct}:
Answer \texttt{yes} if \texttt{extracted\_final\_answer} matches \texttt{[correct\_answer]},
or is within a small margin of error for numerical problems.
Answer \texttt{no} otherwise, i.e., if there is any inconsistency, ambiguity, non-equivalency, or if the extracted answer is incorrect.

\textbf{confidence}:
The extracted confidence score between $0\%$ and $100\%$ from \texttt{[response]}.
Put $100$ if no confidence score is available.
\end{AIbox}
\caption{Prompt for final-answer correctness evaluation.}
\label{fig:answer_grading_prompt}
\end{figure*}

\section{Tool Specifications}

The complete function schemas are shown in
Figures~\ref{fig:tool_search}--\ref{fig:tool_refute}.
The five tools support three stages of the propose--falsify research
process. First, \texttt{search} and \texttt{visit} provide evidence
acquisition. The former executes multiple complementary queries in one
call, while the latter retrieves goal-conditioned evidence from one or
more webpages.

Second, \texttt{update\_context} serves as the central research-state
transition interface. It serializes the current interaction history into
an evidence-grounded context, an explicit constraint list, and a set of
candidate hypotheses. When candidate hypotheses are provided, the runtime
invokes independent Falsifiers and appends their reports to the refreshed
research state, thereby connecting context compression with active
falsification.

Finally, \texttt{finish} and \texttt{refute} define the
terminal actions of the Searcher and Falsifier, respectively. The Searcher
uses \texttt{finish} to return a concise answer together with supporting
evidence and a confidence estimate. The Falsifier uses
\texttt{refute} to return a candidate-level verdict,
constraint status, relevant sources, and recommended follow-up searches.
Together, these tools provide a structured interface for evidence
acquisition, research-state revision, and answer synthesis.

\begin{figure*}[t]
\begin{AIbox}{Tool: search}
\begin{lstlisting}[language={},basicstyle=\ttfamily\footnotesize,breaklines=true,frame=single,columns=fullflexible]
{
  "type": "function",
  "function": {
    "name": "search",
    "description": "Performs batched web searches: supply an array 'query'; the tool retrieves the top 10 results for each query in one call",
    "parameters": {
      "type": "object",
      "properties": {
        "query": {
          "type": "array",
          "items": { "type": "string" },
          "description": "Array of query strings. Include multiple complementary search queries in a single call."
        }
      },
      "required": ["query"]
    }
  }
}
\end{lstlisting}
\end{AIbox}
\caption{Function specification of the \texttt{search} tool.}
\label{fig:tool_search}
\end{figure*}

\begin{figure*}[t]
\begin{AIbox}{Tool: visit}
\begin{lstlisting}[language={},basicstyle=\ttfamily\footnotesize,breaklines=true,frame=single,columns=fullflexible]
{
  "type": "function",
  "function": {
    "name": "visit",
    "description": "Visit webpage(s) and return the summary of the content.",
    "parameters": {
      "type": "object",
      "properties": {
        "url": {
          "type": ["string", "array"],
          "items": { "type": "string" },
          "minItems": 1,
          "description": "The URL(s) of the webpage(s) to visit. Can be a single URL or an array of URLs."
        },
        "goal": {
          "type": "string",
          "description": "The goal of the visit for webpage(s)."
        }
      },
      "required": ["url", "goal"]
    }
  }
}
\end{lstlisting}
\end{AIbox}
\caption{Function specification of the \texttt{visit} tool.}
\label{fig:tool_visit}
\end{figure*}

\begin{figure*}[t]
\begin{AIbox}{Tool: update\_context}
\begin{lstlisting}[language={},basicstyle=\ttfamily\tiny,breaklines=true,frame=single,columns=fullflexible]
{
  "type": "function",
  "function": {
    "name": "update_context",
    "description": "Compress the current research state and trigger the adversarial Falsifier checkpoint. Call this tool only by itself when context is long, the search state is cluttered, or you have reached a meaningful candidate-analysis milestone. `context` must be dense and evidence-grounded. `constraints` and `candidates` must be valid JSON arrays; when candidates is non-empty, independent Falsifier agents attack them and their reports are appended to the refreshed context.",
    "parameters": {
      "type": "object",
      "properties": {
        "context": {
          "type": "string",
          "description": "Dense, lossless research state with confirmed facts, URLs, constraints, candidates, failed paths, and next steps."
        },
        "constraints": {
          "type": "array",
          "description": "Task constraints and their current evidence status.",
          "items": {
            "type": "object",
            "properties": {
              "constraint_id": { "type": "string" },
              "constraint": { "type": "string" },
              "type": { "type": "string", "description": "hard or soft" },
              "status": { "type": "string", "description": "verified, contradicted, unresolved, or unknown" },
              "evidence": { "type": "string" },
              "url": { "type": "string" }
            },
            "required": ["constraint"]
          }
        },
        "candidates": {
          "type": "array",
          "description": "Candidate hypotheses for adversarial falsification. Use [] if none exists yet.",
          "items": {
            "type": "object",
            "properties": {
              "candidate_id": { "type": "string" },
              "answer": { "type": "string" },
              "supporting_evidence": {
                "type": "array",
                "items": {
                  "type": "object",
                  "properties": {
                    "evidence": { "type": "string" },
                    "url": { "type": "string" }
                  }
                }
              },
              "unresolved_constraints": {
                "type": "array",
                "items": { "type": "string" }
              },
              "status": { "type": "string" },
              "notes": { "type": "string" }
            },
            "required": ["answer"]
          }
        }
      },
      "required": ["context", "constraints", "candidates"]
    }
  }
}
\end{lstlisting}
\end{AIbox}
\caption{Function specification of the \texttt{update\_context} tool.}
\label{fig:tool_update_context}
\end{figure*}

\begin{figure*}[t]
\begin{AIbox}{Tool: finish}
\begin{lstlisting}[language={},basicstyle=\ttfamily\footnotesize,breaklines=true,frame=single,columns=fullflexible]
{
  "type": "function",
  "function": {
    "name": "finish",
    "description": "Return the final result when you have a definitive answer. This function signals that your research is complete and you're ready to present the final answer to the user.",
    "parameters": {
      "type": "object",
      "properties": {
        "answer": {
          "type": "string",
          "description": "A succinct, final answer to the user's query"
        },
        "evidences": {
          "type": "array",
          "description": "Array of evidence objects supporting the final answer. Each piece of evidence must cite exactly one url.",
          "items": {
            "type": "object",
            "properties": {
              "evidence": {
                "type": "string",
                "description": "The specific verified fact or data point (one sentence)"
              },
              "url": {
                "type": "string",
                "description": "The source URL of the evidence"
              }
            },
            "required": ["evidence", "url"]
          }
        },
        "confidence": {
          "type": "string",
          "description": "Your confidence score between 0% and 100% for your answer"
        }
      },
      "required": ["answer", "evidences", "confidence"]
    }
  }
}
\end{lstlisting}
\end{AIbox}
\caption{Function specification of the \texttt{finish} tool.}
\label{fig:tool_finish}
\end{figure*}

\begin{figure*}[t]
\begin{AIbox}{Tool: refute}
\begin{lstlisting}[language={},basicstyle=\ttfamily\footnotesize,breaklines=true,frame=single,columns=fullflexible]
{
  "type": "function",
  "function": {
    "name": "refute",
    "description": "Return the final falsification report for the single assigned candidate. This does not answer the original question.",
    "parameters": {
      "type": "object",
      "properties": {
        "candidate_id": { "type": "string" },
        "candidate_answer": { "type": "string" },
        "verdict": { "type": "string", "enum": ["reject", "survive", "uncertain"] },
        "refuted_constraints": { "type": "array", "items": { "type": "object" } },
        "supporting_constraints": { "type": "array", "items": { "type": "object" } },
        "unresolved_constraints": { "type": "array", "items": { "type": "object" } },
        "refuting_sources": { "type": "array", "items": { "type": "object" } },
        "supporting_sources": { "type": "array", "items": { "type": "object" } },
        "recommended_next_searches": { "type": "array", "items": { "type": "string" } },
        "summary": { "type": "string" }
      },
      "required": ["candidate_answer", "verdict", "summary"]
    }
  }
}
\end{lstlisting}
\end{AIbox}
\caption{Function specification of the \texttt{refute} tool.}
\label{fig:tool_refute}
\end{figure*}

\section{Case Studies}

\begin{figure*}
    \centering
    \includegraphics[width=\linewidth]{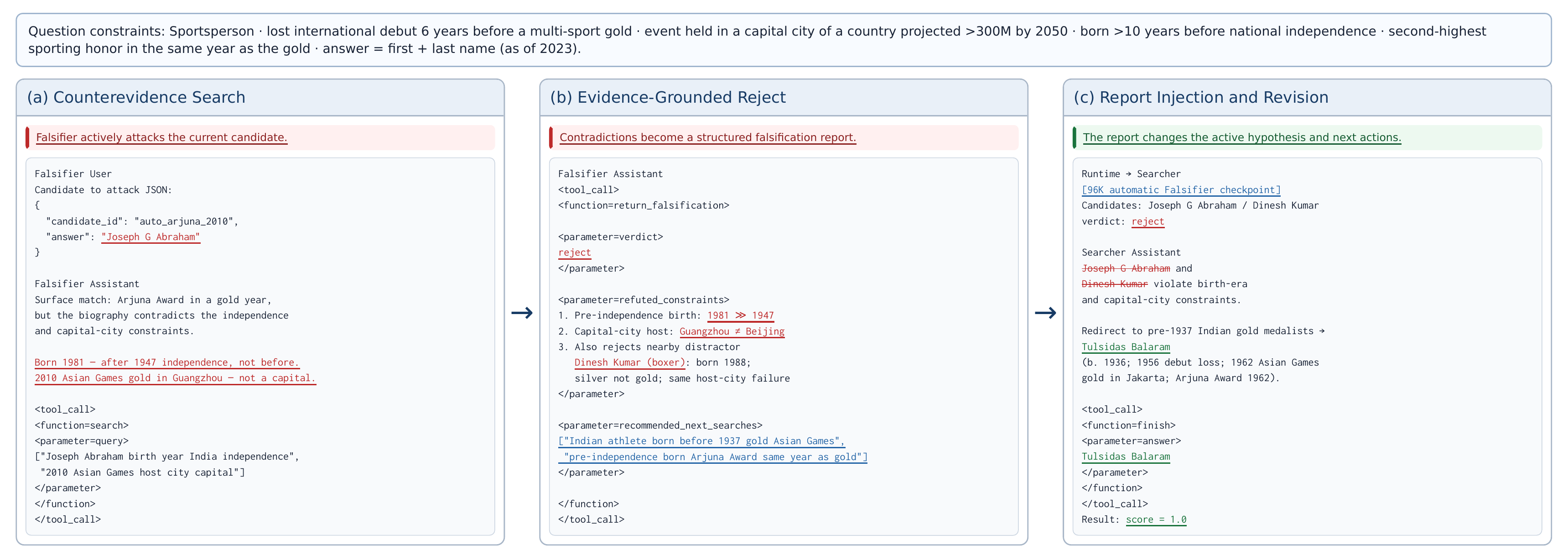}
    \caption{\textbf{Case study of constraint-based candidate rejection.} The Falsifier identifies decisive biographical and geographic contradictions in the initial candidate, returns a structured rejection report, and redirects the Searcher to the correct answer, Tulsidas Balaram.}
    \label{fig:case_tulsidas}
\end{figure*}

\begin{figure*}
    \centering
    \includegraphics[width=\linewidth]{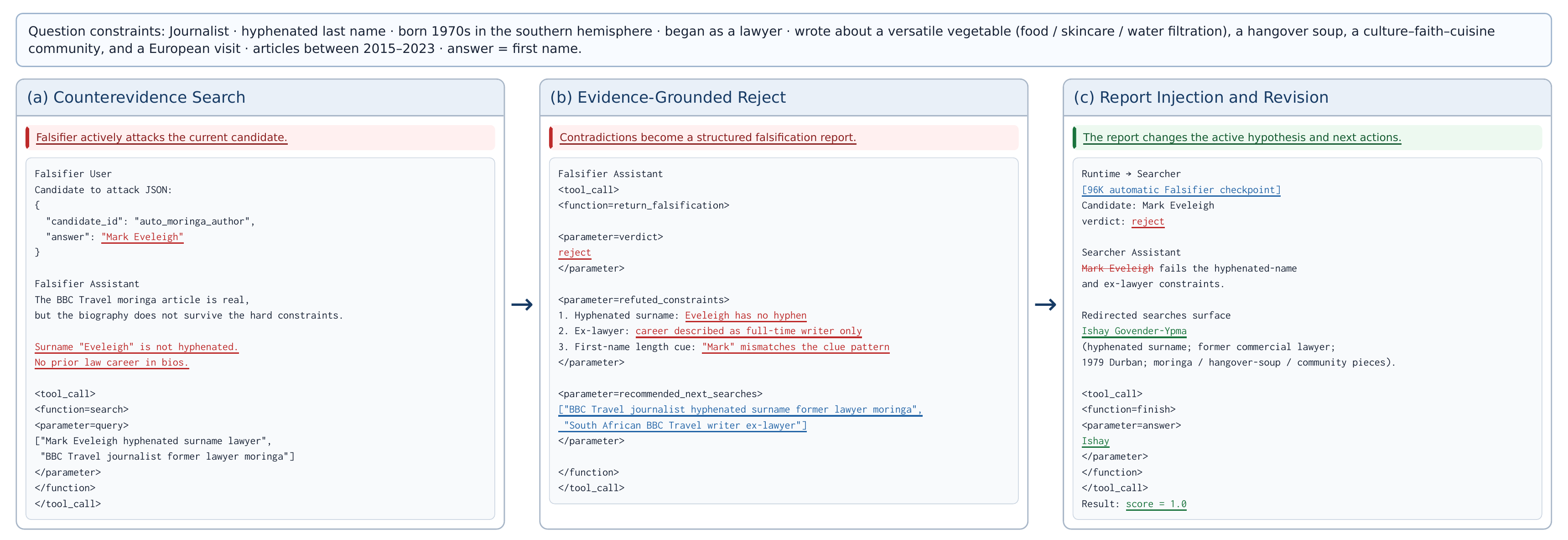}
    \caption{\textbf{Case study of biography-grounded search revision.}The Falsifier rejects a topic-matching but biographically inconsistent candidate and redirects the Searcher to Ishay Govender-Ypma.}
    \label{fig:case_ishay}
\end{figure*}

Figures~\ref{fig:case_tulsidas} and~\ref{fig:case_ishay} illustrate how falsification changes the research trajectory rather than merely verifying the final answer.
Each case follows three stages: the Falsifier first searches for counterevidence against the current candidate, then returns a structured verdict with violated constraints and recommended next searches, and finally injects the report into the Searcher context to trigger hypothesis revision.

Figure~\ref{fig:case_tulsidas} shows a sportsperson-identification task with multiple biographical, temporal, and geographic constraints.
The initial candidate, Joseph G Abraham, superficially matches the award and gold-medal clues, but violates two decisive conditions: he was born after Indian independence, and his gold-medal event was held in Guangzhou rather than a capital city.
The Falsifier converts these contradictions into a structured rejection report and recommends searches centered on pre-independence-born Indian gold medalists.
After receiving the report, the Searcher discards both Joseph G Abraham and a nearby distractor, redirects the search, and identifies Tulsidas Balaram as the verified answer.

Figure~\ref{fig:case_ishay} presents an author-identification task in which the initial candidate, Mark Eveleigh, is strongly associated with the relevant article topic but fails the biographical constraints.
In particular, his surname is not hyphenated and available biographies do not support a prior legal career.
The Falsifier therefore rejects the candidate and proposes more discriminative searches combining the publication, hyphenated surname, and former-lawyer attributes.
The resulting context revision leads the Searcher to Ishay Govender-Ypma, after which the Searcher returns the required first name, Ishay.

\end{document}